\documentclass[sigconf, screen]{acmart}

\renewcommand\footnotetextcopyrightpermission[1]{} 
\AtBeginDocument{%
  \providecommand\BibTeX{{%
    Bib\TeX}}}

\setcopyright{acmlicensed}
\copyrightyear{2026}
\acmYear{2026}
\acmDOI{XXXXXXX.XXXXXXX}
\acmConference[]{}
\acmISBN{}

\usepackage{multirow}
\usepackage{booktabs}   
\usepackage{graphicx}   
\usepackage{caption}    
\usepackage{natbib}
\usepackage{subcaption} 
\usepackage{array}
\usepackage{colortbl}
\usepackage{longtable}

\definecolor{lightb}{RGB}{225,240,255}
\definecolor{darkb}{RGB}{190,220,255}
\definecolor{grayg}{RGB}{210, 210, 210}   
\definecolor{lightg}{RGB}{235, 235, 235}

\usepackage{multirow}
\usepackage{graphicx}
\usepackage{float}
\usepackage{xcolor}
\usepackage{booktabs}
\usepackage{array}
\usepackage{tabularx}
\usepackage{makecell}
\usepackage{soul}
\usepackage{CJKutf8}

\definecolor{hlgreen}{RGB}{198,239,206}
\definecolor{hlyellow}{RGB}{255,235,156}
\definecolor{hlpurple}{RGB}{220,198,224}
\definecolor{hlred}{RGB}{255,199,206}

\newcommand{\hgreen}[1]{\colorbox{hlgreen}{\begin{CJK}{UTF8}{gbsn}#1\end{CJK}}}
\newcommand{\hyellow}[1]{\colorbox{hlyellow}{\begin{CJK}{UTF8}{gbsn}#1\end{CJK}}}
\newcommand{\hpurple}[1]{\colorbox{hlpurple}{\begin{CJK}{UTF8}{gbsn}#1\end{CJK}}}
\newcommand{\hred}[1]{\colorbox{hlred}{\begin{CJK}{UTF8}{gbsn}#1\end{CJK}}}

\newcommand{\hgreene}[1]{{\sethlcolor{hlgreen}\hl{#1}}}
\newcommand{\hyellowe}[1]{{\sethlcolor{hlyellow}\hl{#1}}}
\newcommand{\hpurplee}[1]{{\sethlcolor{hlpurple}\hl{#1}}}
\newcommand{\hrede}[1]{{\sethlcolor{hlred}\hl{#1}}}

\usepackage[labelfont=bf,skip=0pt]{caption}
\AtBeginDocument{%
  \providecommand\BibTeX{{%
    \normalfont B\kern-0.5em{\scshape i\kern-0.25em b}\kern-0.8em\TeX}}}

\renewcommand\footnotetextcopyrightpermission[1]{} 
\acmSubmissionID{4537}

\begin{document}

\title{SignDPO: Multi-level Direct Preference Optimisation for \\ Skeleton-based Gloss-free Sign Language Translation}


\author{Muxin Pu$^{1,2}$, Xiao-Ming Wu$^2$, Mei Kuan Lim$^1$, Chun Yong Chong$^1$, Wei Li$^2$, Chen Change Loy$^2$}
\affiliation{%
  \institution{$^1$Monash University, $^2$Nanyang Technological University}
  \country{} 
}
\email{{muxin.pu@monash.edu}}

\renewcommand{\shortauthors}{Pu et al.}

\begin{abstract}

We present SignDPO, a novel multi-level Direct Preference Optimisation (DPO) framework designed to enhance the alignment of skeleton-based Sign Language Translation. While current skeleton-based models have made significant progress using Maximum Likelihood Estimation, they are primarily constrained by an imitation-based paradigm that lacks discriminative sensitivity to the fine-grained spatio-temporal nuances of sign language, often leading to semantic drift. To address this, SignDPO shifts the optimisation goal from simple sequence mimicry to structured preference alignment across spatial, temporal, and linguistic dimensions. Our framework involves three key designs. First, we introduce a hierarchical perturbation strategy to construct spatial and temporal non-preferred samples at both global and local granularities automatically. Second, we propose a self-guiding mechanism that leverages decoder cross-attention scores to identify and perturb semantically salient skeletal regions, forcing the model to distinguish genuine sign signals from structural distortions. Third, we establish an automated language-level preference generator by fine-tuning a dedicated perturbation model, capturing complex output-level failure modes without manual annotation. Extensive experiments on three widely adopted benchmarks, CSL-Daily, How2Sign, and OpenASL, demonstrate that SignDPO consistently outperforms state-of-the-art gloss-free methods and even rivals established gloss-based ones. Our results suggest that multi-level preference alignment is a powerful paradigm for bridging the gap between high-entropy skeletal trajectories and discrete linguistic semantics. The artefacts and code related to this study are made publicly online. \footnote{https://github.com/mpuu00001/SignDPO}

\end{abstract}





\captionsetup[figure]{skip=5.5pt} 
\captionsetup[table]{skip=5.5pt}   
\setlength{\abovecaptionskip}{5.5pt}   
\setlength{\belowcaptionskip}{5.5pt}   

\setlength{\textfloatsep}{5.5pt} 
\setlength{\intextsep}{5.5pt} 
\setlength{\floatsep}{5.5pt} 



\maketitle

\begin{figure}[!t]
  \centering
  \includegraphics[scale=0.42]{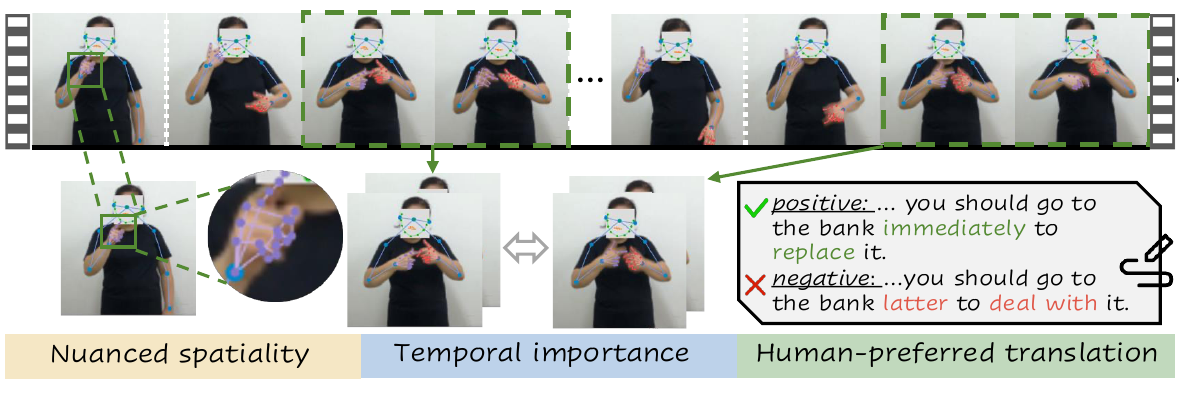}
  \caption{Challenges in capturing the multi-dimensional semantics of sign language. SLT requires modelling fine-grained spatial and temporal semantics, yet existing methods often overlook these cues, leading to lexically correct but semantically deficient translations.}
  \label{fig:questions}
\end{figure}

\section{Introduction}
Sign language serves as a vital linguistic bridge for the community of people with speech or hearing impairments \citep{WHO, WFD}, embodying a rich, complex, multidimensional communication system \citep{sandler2006sign}. In recent years, Sign Language Translation (SLT) has shifted from traditional gloss-based pipelines \citep{camgoz2020sign, zhou2021improving, zhang2023sltunet, chen2022two, zhao2024conditional} that rely on costly and often ambiguous intermediate annotations to end-to-end gloss-free frameworks \citep{yin2023gloss, zhou2024scaling, zhou2023gloss, chen2025c, wong2024sign2gpt}. Within this transition, skeleton-based approaches \cite{gan2023towards, Li2025sign, pu2025sigma} have emerged as a trend due to their computational efficiency, robustness to environmental noise, and inherent privacy-preserving nature. By modelling body keypoints with graph neural networks and mapping them to natural language using pre-trained models \cite{Li2025sign, pu2025sigma}, these approaches have achieved state-of-the-art performance in cross-modal sign-to-text generation.

The inherent challenge of skeleton-based SLT lies in the intricate mapping between high-entropy spatio-temporal trajectories and discrete linguistic semantics \citep{camgoz2018neural}. Unlike general action recognition \citep{yan2018spatial}, sign language conveys meaning through a highly dense and fine-grained coordination of movements \citep{sandler2006sign}. As illustrated in Figure \ref{fig:questions}, in the spatial dimension, a minute shift in hand orientation or a subtle change in finger configuration can pivot the entire semantic meaning of a sentence. Simultaneously, in the temporal dimension, the velocity, rhythm, and sequential order of gestures encode critical grammatical markers that are absent in static frames.

To resolve these intricate spatio-temporal dependencies, current skeleton-based systems are predominantly optimised via Maximum Likelihood Estimation (MLE) \citep{zhou2024scaling, Li2025sign, pu2025sigma}, which treats translation as a mimicry of ground-truth sequences. However, we argue that this imitation-based paradigm is inherently limited in fulfilling such fine-grained requirements for two reasons. First, MLE lacks a discriminative training signal to distinguish semantically decisive cues from redundant motion noise; it provides no explicit mechanism to penalise plausible-but-incorrect translations \citep{ouyang2022training, rafailov2023direct} (e.g., the negative sample in Figure \ref{fig:questions}, rightmost). This leads to frequent semantic drift, where the output remains linguistically fluent but factually decoupled from the source gesture. Second, the teacher-forcing nature of MLE makes models insensitive to structural perturbations. Without a contrastive signal to highlight how spatial or temporal distortions alter meaning, models often rely on language priors rather than genuinely grounding their translations in the visual evidence. 

To bridge this gap, we take inspiration from the recent success of human preference alignment in large language models \citep{ouyang2022training, 
christiano2017deep}. Direct Preference Optimisation (DPO) has emerged as a powerful alternative to reinforcement learning \citep{rafailov2023direct}, allowing models to learn from contrastive pairs of preferred and non-preferred responses. While DPO has been successfully applied to general vision-language tasks \citep{zhang2025direct, huang2025vistadpo}, its application to sign language remains non-trivial. Unlike static images or general videos, sign language carries dense, structured semantic content embedded within skeletal trajectories \citep{pu2025sigma, Li2025sign}. A naive application of DPO at the text level is insufficient to capture the modality-specific failure modes inherent in SLT.

\begin{figure}[thbp]
  \centering
  \includegraphics[scale=1]{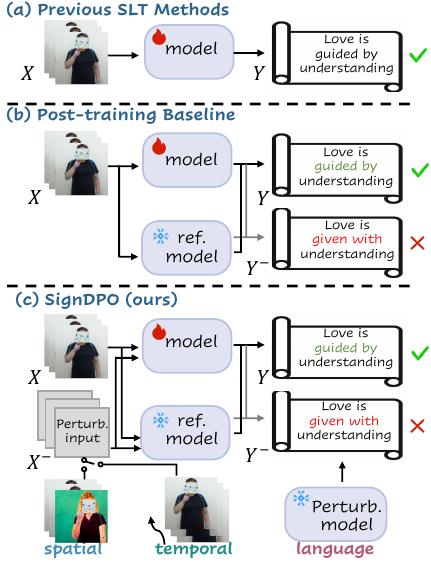}
  \caption{The overview illustrates the difference between (a) previous SLT methods, (b) the post-training baseline (standard DPO), and (c) our method SignDPO.}
  \label{fig:compare}
\end{figure}

In this paper, we introduce SignDPO, a novel multi-level DPO framework designed for skeleton-based SLT. Figure \ref{fig:compare} compares our work to previous SLT methods and the post-training baseline (standard DPO). Our core insight is that for a sign language model to be aligned, it must be sensitive to preferences across three distinct dimensions: spatial configurations, temporal dynamics, and linguistic fluency. To achieve this, we propose a hierarchical perturbation strategy to automatically construct non-preferred (negative) samples. At the spatial and temporal level, we construct non-preferred samples by applying perturbations at both the global full-frame level and within local salient regions. The latter utilises a self-guiding mechanism, leveraging cross-attention scores from the decoder, to identify semantically critical key frames. This multi-granularity approach forces the model to distinguish between genuine sign signals and corrupted inputs. At the language level, we fine-tune a specialised perturbation model to generate partially semantically adjacent but inaccurate translations, providing a high-quality contrastive signal for linguistic alignment.

By incorporating multi-level preference signals into a unified objective, SignDPO encourages the model to go beyond simple sequence imitation and produce text that more accurately reflects the underlying spatio-temporal structure of sign language. We evaluate SignDPO on three widely-used benchmarks: CSL-Daily \citep{zhou2021improving}, How2Sign \citep{duarte2021how2sign}, and OpenASL \citep{shi2022open}. Experimental results demonstrate that SignDPO consistently outperforms state-of-the-art gloss-free methods and even rivals established gloss-based systems. Our main contributions are threefold:

\begin{itemize}
    \item We propose SignDPO, the first multi-level preference optimization framework for SLT, which aligns models with human-like sensitivity to spatial, temporal, and linguistic nuances.
    \item We introduce a hierarchical perturbation strategy that utilizes cross-attention importance scores to adaptively generate negative samples, enabling more targeted and effective contrastive learning in the visual domain.
    \item We establish an automated language-level preference generator by fine-tuning a dedicated perturbation model, bypassing the need for manual negative sample annotation while capturing complex output-level failure modes.
\end{itemize}

Our method achieves state-of-the-art performance across three major SLT benchmarks (CSL-Daily, How2Sign, OpenASL), demonstrating the superior generalization ability of preference-based alignment in sign language.

\section{Related works}
\subsection{Sign Language Translation}
SLT has evolved from gloss-based pipelines, which decompose the task into recognition and then translation \citep{camgoz2018neural, zhou2021improving}, to end-to-end models that directly map visual input to natural language \citep{camgoz2020sign}. While glosses provide intermediate supervision, they only approximate semantic content, are costly to annotate, and introduce a semantic bottleneck. The shift toward gloss-free SLT has therefore been driven by advances in pre-trained language models and improved representation alignment.

The field of SLT has shifted from multi-stage pipelines toward a gloss-free paradigm, an approach pioneered by showing that end-to-end visual-language training can match gloss-based systems \citep{camgoz2020sign}. To bridge the modality gap, transfer learning has been employed to progressively pre-train visual and language modules before integrating them for joint translation \citep{chen2022simple}. Subsequent research has largely focused on leveraging contrastive learning to strengthen cross-modal alignment. For instance, token-level objectives have been used to prevent representation collapse \citep{fu2023conslt}, while dual visual and semantic losses \citep{gan2023contrastive} and weakly labelled data \citep{zhou2023gloss} have been utilised to enhance pre-training. Self-supervised contrastive strategies further refine representations by separating clustered gesture features to mitigate density bottlenecks \citep{ye2024improving}. Hierarchical global and local contrastive objectives have been unified with sign-aware fusion into single pre-training frameworks \citep{pu2025sigma}. A complementary research direction has explored the adaptation of large language models through sign-text fine-tuning, directly exploiting their powerful generative priors for translation \citep{wong2024sign2gpt}. More recently, RVLF \citep{rao2025rvlf} applies Group Relative Policy Optimisation \cite{shao2024deepseekmath} with sentence-level rewards, yet remains confined to output-level alignment without addressing the spatio-temporal perceptual failures motivating SignDPO.

Parallel Skeleton-based approaches offer computational efficiency and robustness \cite{yan2018spatial}, with models like SignBERT+ \cite{hu2023signbertplus}, Uni-sign \cite{Li2025sign}, and Sigma \cite{pu2025sigma} pre-training on large-scale sequences for general sign representations. Our backbone follows this paradigm, yet unlike prior methods that rely on MLE for fine-tuning, SignDPO further addresses its inability to penalise plausible-but-incorrect translations via preference-based supervision across spatial, temporal, and language dimensions.

\subsection{Preference Optimisation and Alignment}
The alignment of generative models with human preferences has traditionally been addressed through Reinforcement Learning from Human Feedback (RLHF), where a reward model trained on human judgments guides policy optimisation via Proximal Policy Optimisation (PPO) \citep{christiano2017deep, ouyang2022training}. To simplify this, Direct Preference Optimisation (DPO) was introduced as a more stable alternative that optimises directly over preferred and rejected pairs without an explicit reward model \citep{rafailov2023direct}, and has since been adopted across instruction following \citep{touvron2023llama2}, summarisation \citep{stiennon2020learning}, and code generation \citep{zhang2025focused}.

Subsequent extensions of DPO have explored richer preference structures and automated data generation. Generalised frameworks have been proposed to unify several preference optimisation objectives \citep{azar2024general}, while self-play mechanisms now allow for the generation of preference data without human annotation \citep{chen2024self}. Within the domain of vision-language models, DPO-style training has been successfully applied to visual question answering \citep{li2023silkie} and extended to complex video-understanding tasks \citep{zhou2024aligning, zhang2025direct}. SignDPO is inspired by this preference-based paradigm, extending it to skeleton-based SLT, where dense spatio-temporal semantics and different failure modes demand a more structured preference signal that goes beyond standard video-language alignment frameworks.

\begin{figure*}[thbp!]
  \centering
  \includegraphics[width=1\textwidth]{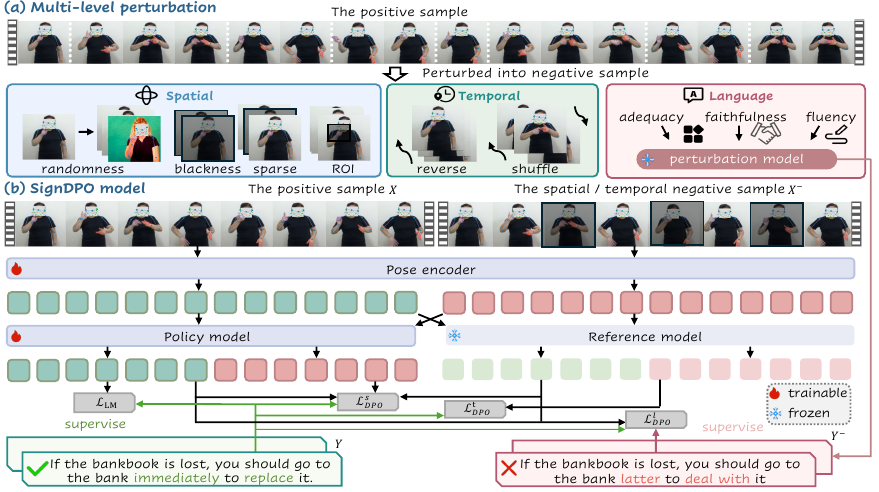}
  \caption{Overview of the SignDPO Framework. (a) Multi-level perturbation: Non-preferred (negative) samples are constructed across three dimensions across spatial, temporal, and linguistic dimensions using global/local strategies (e.g., randomness, blackness, sparse masks, ROI masks, reverse, and shuffle) and a score-conditioned (adequacy, faithfulness, and fluency) language perturbation model. (b) SignDPO model: The policy model $\pi_\theta$ is trained to maximise the log-likelihood gap between the preferred (positive) sample $X$ and the perturbed negative samples $X^-$. This process is regularised by a frozen reference model $\pi_{\text{ref}}$ and a joint objective that aligns the model across spatial ($\mathcal{L}_{\text{DPO}}^s$), temporal ($\mathcal{L}_{\text{DPO}}^t$), and linguistic ($\mathcal{L}_{\text{DPO}}^l$ + $\mathcal{L}_{LM}$ ) levels.}
  \label{fig:questions}
\end{figure*}


\section{Methodology}

\subsection{Problem Formulation}
We consider SLT a sequence-to-sequence generation task. Given a skeletal sequence $X = \{x_1, x_2, \ldots, x_T\}$, where each frame $x_t \in \mathbb{R}^{69 \times 3}$ encodes the $(x, y, \text{confidence})$ coordinates of 69 body keypoints spanning the hands (21 keypoints per hand), face (18 keypoints), and upper body (9 keypoints), the model is trained to generate a natural language translation $Y = \{y_1, y_2, \ldots, y_L\}$.

Standard training maximises the likelihood $p(Y \mid X)$ via a cross-entropy objective. However, this provides only positive supervision, offering no signal regarding which input features to prioritise or which outputs to avoid. To address this, we build on DPO \cite{rafailov2023direct}. Given a policy $\pi_\theta$, a frozen reference model $\pi_{\text{ref}}$, and a preference pair $(Y, Y^-)$, DPO optimises the following objective:

\begin{equation}
    u(X, Y, Y^-) = \beta \log \frac{\pi_\theta(Y \mid X)}{\pi_{\text{ref}}(Y \mid X)} - \beta \log \frac{\pi_\theta(Y^- \mid X)}{\pi_{\text{ref}}(Y^- \mid X)}
\end{equation}
\begin{equation}
    \mathcal{L}_{\text{DPO}}(X, Y, Y^-) = -\mathbb{E}_{(X, Y, Y^-)}\left[\log \sigma \left( u(X, Y, Y^-) \right)\right]
\end{equation}
where $\beta$ controls the deviation from the reference policy, and the sequence log-likelihood can be formulated as:

\begin{equation}
    \log \pi(Y \mid X) = \sum_{y_i \in Y} \log p(y_i \mid X, y_{<i})
\end{equation}

Crucially, we extend the standard DPO formulation to the multimodal setting of SLT. Rather than constructing non-preferred samples exclusively at the output level. SignDPO constructs non-preferred samples by applying structured perturbations to the visual input $X$, yielding perturbed inputs $X^-$ that degrade the spatial or temporal integrity of the skeleton sequence. This allows preference signals to be defined at the input level, directly targeting the spatial and temporal failure modes of SLT models. Language-level non-preferred samples $Y^-$, generated by a fine-tuned perturbation model, further capture output-level preference misalignment. 

\subsection{Backbone}
Our backbone follows a skeleton-based encoder-decoder architecture. The visual encoder consists of part-specific Spatial-Temporal Graph Convolutional Networks (ST-GCN) \cite{yan2018spatial} that model each body part, left hand, right hand, body, and face, independently. For each part $p$, the raw skeletal input $S_p^{\text{raw}} \in \mathbb{R}^{L \times N_p \times 3}$ is projected to a compact feature $S_p \in \mathbb{R}^{L \times D}$, then processed through a spatial GCN that captures joint interdependencies and a temporal GCN that models motion dynamics. The outputs of all four parts are concatenated and projected via a pose projection layer to produce the visual representation $S \in \mathbb{R}^{L \times D_{\text{model}}}$, which is then prepended with a language prefix (prompt) and fed into a pre-trained mT5-base \cite{xue2020mt5} to generate the target translation. During training, the policy model $\pi_\theta$ is initialised from a pre-trained checkpoint (explained in Section \ref{sec:exp}) with its mT5 parameters fully fine-tuned, and the reference model $\pi_{\text{ref}}$ remains a frozen copy of this initial state, consistent with the standard DPO framework.

\subsection{Spatial-level DPO}
The spatial arrangement of keypoints in sign language conveys direct semantic meaning. However, standard cross-entropy training lacks a mechanism to penalise the model for ignoring or misinterpreting spatially informative regions. We address this through spatial-level preference optimisation, in which non-preferred samples are constructed by degrading the spatial integrity of the input.

\paragraph{Global spatial perturbations.} We define four strategies that corrupt the spatial structure uniformly across all frames:
\begin{itemize}
    \item \textbf{Randomness}: replaces the skeletal sequence of a given sample with that of a randomly selected sample from the same mini-batch, introducing a fully mismatched spatial configuration while preserving temporal coherence.
    \item \textbf{Blackness}: sets all keypoint coordinates to zero across every frame, eliminating the spatial signal entirely.
    \item \textbf{Random Sparse Mask}: randomly and sparsely zeros out a subset of frames, with the constraint that masked frames are non-contiguous, preventing the model from learning positional shortcuts.
    \item \textbf{ROI Mask}: removes the semantically critical spatial region by zeroing out the dominant hand. During training, the dominant hand is sampled "on-the-fly" with a $75\%$ probability for the right hand and $25\%$ for the left, targeting the core gestural signals.
\end{itemize}

\paragraph{Local spatial perturbations.} While global perturbations target the sequence as a whole, local perturbations concentrate degradation on the frames that carry the highest semantic weight, providing a more targeted contrastive signal. Key frames are identified adaptively using the cross-attention scores produced by the mT5 decoder during the forward pass on the clean input. Let $\{\mathbf{C}_l\}_{l=1}^{L}$ denote the cross-attention tensors across $L$ decoder layers, where each $\mathbf{C}_l \in \mathbb{R}^{B \times H \times T_{\text{tgt}} \times T_{\text{src}}}$, with $H$ attention heads, $T_{\text{tgt}}$ decoding steps, and $T_{\text{src}} = P + T$ source positions comprising a language prefix of length $P$ followed by $T$ frame tokens. We average over layers, heads, and decoding steps to obtain per-source-position importance scores $\mathbf{a} \in \mathbb{R}^{B \times T_{\text{src}}}$, then discard the prefix positions to retain the frame-level scores $\mathbf{s} \in \mathbb{R}^{B \times T}$:

\begin{equation}
    \mathbf{a} = \frac{1}{L \cdot H \cdot T_{\text{tgt}}} \sum_{l=1}^{L} \sum_{h=1}^{H} \sum_{j=1}^{T_{\text{tgt}}} \mathbf{C}_{l,h,j,:},
    \mathbf{s}_t = \mathbf{a}_{:, P}
\end{equation}

where $t = 1, \ldots, T$. The key frame $k^*$ for each sample is then identified as the frame with the maximum attention weight $k^* = \arg\max_{t} \mathbf{s}_t$. A local window $\mathcal{W}$ of proportional width $wT$ is constructed symmetrically around $k^*$:

\begin{equation}
    \mathcal{W} = \left[\max\left(1, k^* - \left\lfloor \frac{wT}{2} \right\rfloor\right), \min\left(T, k^* + \left\lfloor \frac{wT}{2} \right\rfloor\right)\right]
\end{equation}

where $w \in (0, 1)$ is the window ratio hyperparameter. The perturbation strategies are then applied exclusively within $\mathcal{W}$, leaving frames outside the window unchanged.

\paragraph{Spatial DPO loss.} The spatial loss integrates global and local non-preferred samples, where one of four strategies is randomly chosen with equal probability at each granularity during training.
\begin{equation} \label{eq:sp_loss}
    \mathcal{L}_{\text{DPO}}^s(X, Y, X_{\text{s-}}^-, X_{\text{s-l}}^-) = \frac{1}{2}[\mathcal{L}_{\text{DPO}}^{*}(X, Y, X_{\text{s-g}}^-) + \mathcal{L}_{\text{DPO}}^{*}(X, Y, X_{\text{s-l}}^-)]
\end{equation}
where
\begin{equation}
    \mathcal{L}^{*}_{\text{DPO}}(X, Y, X^-) = -\mathbb{E}_{(X, Y, X^-)}\left[\log \sigma \left( u^{*}(X, Y, X^-) \right)\right]
\end{equation}
where
\begin{equation}
    u^{*}(X, Y, X^-) = \log \frac{\pi_{\theta}(Y \mid X)}{\pi_{\text{ref}}(Y \mid X)} - \log \frac{\pi_{\theta}(Y \mid X^-)}{\pi_{\text{ref}}(Y \mid X^-)}
\end{equation}


\begin{figure*}[thbp!]
  \centering
  \includegraphics[scale=1]{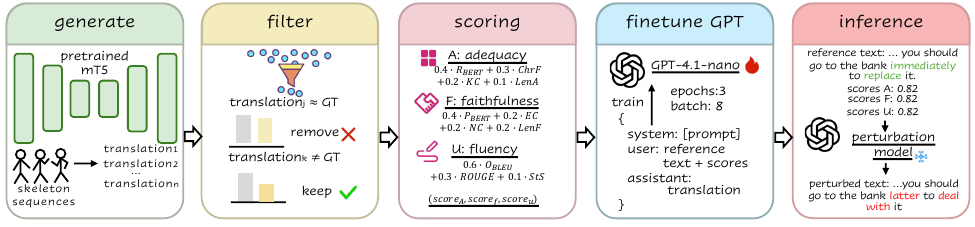}
  \caption{Pipeline of the Language Perturbation Model. Imperfect translations are first generated by the pre-trained SLT model and filtered against ground-truth references. These samples are then automatically scored based on semantic adequacy, faithfulness, and linguistic fluency. These reference-score-translation pairs are used to fine-tune a GPT-4.1-nano model, which subsequently acts as an automated generator of semantically adjacent but non-preferred translations ($Y^-$) for our language-level objective.}
  \label{fig:lang-pert}
\end{figure*}

\subsection{Temporal-level DPO}
The temporal ordering of frames in sign languages is semantically meaningful. A model trained with cross-entropy may exploit static frame-level cues or language model priors rather than genuinely learning the temporal dynamics of the gesture sequence. We introduce temporal-level preference optimisation to provide an explicit training signal that rewards temporal understanding.

\paragraph{Global temporal perturbations.} Two strategies disrupt the temporal structure of the full sequence:
\begin{itemize}
    \item \textbf{Reverse}: reverses the temporal order of all frames, producing a sequence in which every gesture unfolds backwards. This directly tests whether the model has learned the directional semantics of sign movements.
    \item \textbf{Shuffle}: randomly permutes all frames, destroying temporal coherence while preserving the spatial content of each individual frame.
\end{itemize}

\paragraph{Local temporal perturbations.} Analogously to the spatial case, local temporal perturbations apply reverse and shuffle operations within the key frame window $\mathcal{W}$ identified via cross-attention scores, leaving frames outside the window in their original order. 

\paragraph{Temporal DPO loss.} The temporal loss integrates global and local non-preferred samples, where either reverse or shuffle is randomly chosen with equal probability at each granularity. Because of the lesser number of perturbation strategies designed for this level compared to the spatial level, both global and local non-preferred samples are incorporated jointly within a single DPO objective. The non-preferred log-ratio term is replaced by a combination of the global and local components:
\begin{equation} \label{eq:tmp_loss}
    \mathcal{L}_{\text{DPO}}^t = -\mathbb{E}_{(X, Y, X_g^-, X_l^-)}\left[\log \sigma\left(u_{\theta}^{\text{combo}} (X, Y, X_g^-, X_l^-)\right)\right]
\end{equation}
where:
\begin{equation}
\begin{split}
    u_{\theta}^{\text{combo}}&(X, Y, X_{g-t}^-, X_{l-t}^-) = \beta \log \frac{\pi_\theta(Y \mid X)}{\pi_{\text{ref}}(Y \mid X)} \\
    &- \frac{1}{2} \sum_{i \in \{\text{g, l}\}} \log \frac{\pi_\theta(Y \mid X_{i-t}^-)}{\pi_{\text{ref}}(Y \mid X_{i-t}^-)}
\end{split}
\end{equation}

\subsection{Language-level DPO}
The spatial and temporal perturbations above teach the model to attend to the correct input structure. However, SLT models can also fail at the output level: they may generate translations that are semantically adjacent to the ground truth yet deviate from natural human expression. These output-level failure modes may fail to be captured by input perturbations. We address them through a dedicated perturbation model (see Figure \ref{fig:lang-pert}) that generates controlled non-preferred translations conditioned on quantified quality deficiencies.

\paragraph{Perturbation model training.} Starting from our pre-trained SLT model, we run inference over the development and testing sets to collect the raw output set $\mathcal{T}_{\text{raw}}$. Outputs identical to the ground truth or reference $X$ are filtered out. Each remaining translation $Y_l$ is then scored along three quality dimensions, where the decimal values in parentheses denote the weight assigned to each sub-metric:

\begin{itemize}
\label{lst:scores}
    \item \textbf{Adequacy $A$}: measures semantic preservation, computed with metrics including BERTScore Recall (0.4), ChrF (0.3), keyword coverage (0.2), and length adequacy (0.1).
    \item \textbf{Faithfulness $F$}: measures hallucination avoidance, got with metrics including BERTScore precision (0.4), combined with entity, number, and length consistency (0.2 each).
    \item \textbf{Fluency $U$}: measures naturalness, computed with metrics including BLEU1-4 (0.6), ROUGE-L (0.3), and structural completeness (0.1).
\end{itemize}

The adequacy ($A$) and faithfulness ($F$) scores are computed as a weighted sum of their respective sub-metrics, expressed as $A = \sum_{i=1}^{4} w_i \cdot \text{metric}_i$ and $F = \sum_{i=1}^{4} w_j \cdot \text{metric}_j$, where the specific weights $w$ are indicated in brackets of the bulleted List \ref{lst:scores}. 

The fluency ($U$) score first computes an overall BLEU score $O_{\text{BLEU}} = \exp\left(\frac{1}{4}\sum_{n=1}^{4} \ln(\text{BLEU}n)\right)$ as the geometric mean of BLEU1-4, penalising translations that fail to balance high unigram coverage with precise higher-order phrase matching. The final fluency is then calculated as a weighted sum $U = \sum_{k=1}^{3} w_k \cdot \text{metric}_k$, where $\text{metric}_k \in \{O_{\text{BLEU}},\, \text{ROUGE-L},$ $\text{structural } \text{completeness}\}$ with weights $w = \{0.6, 0.3, 0.1\}$ respectively.

The scored dataset $\mathcal{D} = \{(X, A, F, U, Y_l)\}$ is used to fine-tune GPT-4.1-nano via supervised fine-tuning. The model learns to generate $Y_l^-$ given a reference translation and three quality scores, providing a scalable and annotation-free generator of language-level non-preferred samples. 

\paragraph{Perturbation model inference.} Linguistic negatives $Y^-_l$ are generated by querying the perturbation model with reference translations and $(A, F, U)$ triples randomly generated.

\paragraph{Language DPO loss.} The language level loss can be expressed as:

\begin{equation}
    \mathcal{L}_{\text{DPO}}^l = \mathcal{L}_{\text{DPO}}(X, Y, Y^-_l)
\end{equation}

\subsection{Joint Training Objective}
SignDPO combines the three preference signals with a language modelling regularisation term into a unified training objective:
\begin{equation} \label{eq:joint_loss}
    \mathcal{L}_{\text{joint}} = \alpha \mathcal{L}_{\text{DPO}}^s + \rho \mathcal{L}_{\text{DPO}}^t + \lambda \frac{1}{2} (\mathcal{L}_{\text{DPO}}^l + \mathcal{L}_{\text{LM}})
\end{equation}
where $\mathcal{L}_{\text{LM}} = -\log \pi_\theta(Y \mid X)$ is the standard cross-entropy loss on the preferred translation, and $\alpha, \rho, \lambda \in \mathbb{R}^+$ are weighting hyperparameters. Spatial and temporal components use cross-attention scores from the preferred output to guide local perturbation windows, creating a self-reinforcing dynamic where accurate localisation produces precisely targeted preference signals.

\section{Experiments} \label{sec:exp}
\textbf{Datasets}. We evaluate SignDPO on three widely adopted benchmarks, including CSL-Daily \citep{zhou2021improving} for Chinese Sign Language (CSL), OpenASL \citep{shi2022open}, and How2Sign \citep{duarte2021how2sign} for American Sign Language (ASL). To obtain a pre-trained SLT model, we follow \cite{pu2025sigma} exactly and exclude the fine-tuning stage, using CSL-News \cite{Li2025sign} for CSL and YouTube-ASL \cite{uthus2023youtube} for ASL, ensuring that the pre-training data are distinct from the downstream evaluation benchmarks. 

\textbf{Evaluation metrics}. Following prior works, we report BLEU \& ROUGE-L scores. For brevity, we denote BLEU-1, BLEU-4, and ROUGE-L as B@1, B@4, and R@L in the tables or figures of the following sections. 

\textbf{Training details}.
Our visual backbone utilises part-specific ST-GCN to extract features from skeleton data. These features are projected into a 768-dimensional latent space and fed into a pre-trained mT5-base model, which serves as our core language generator. The choice of mT5-base allows for robust multilingual support across the Chinese (CSL-Daily) and American (How2Sign, OpenASL) datasets used in our study. We perform preference-based alignment directly on the mT5 model after it is pretrained on CSL-News and YouTube-ASL. We employ the AdamW optimiser with $\beta_1=0.9$, $\beta_2=0.999$, and a weight decay of 1.00E-04. Preference-based alignment is performed with a learning rate of 3.00E-04, governed by a cosine annealing scheduler over 10 epochs. To compute the DPO log-ratios, we maintain a frozen reference model, a deep copy of the model parameters of the pre-trained model, to provide a stable baseline for preference gradients. To manage the memory requirements of multi-modal preference pairs, we utilise the DeepSpeed library with ZeRO Stage 3 optimisation. This setup enables parameter sharding and CPU offloading of optimiser states, allowing us to maintain a total batch size of 10 per GPU in BFloat16 precision across four NVIDIA A100 (80GB) GPUs. The DPO KL-divergence constraint $\beta$ is set to 0.1, utilising a sigmoid loss function with a label smoothing factor of 0.2. For inference, we use beam search with a width of 4 and a maximum length of 100 tokens, selecting the model checkpoint that achieves the highest B@4 score on the development set.

\subsection{Comparison with state-of-the-art methods}
Tables \ref{tab:csl_result}--\ref{tab:openasl_result} compare SignDPO against methods across all three benchmarks, covering both gloss-based and gloss-free paradigms.
 
\textbf{CSL-Daily.} As shown in Table~\ref{tab:csl_result}, SignDPO achieves \textbf{57.43} B@1, \textbf{28.36} B@4, and \textbf{58.55} R@L on the test set, surpassing all prior gloss-free methods. Compared with the strongest gloss-free baseline, Uni-Sign \citep{Li2025sign}, SignDPO improves B@4 by 2.00 points (28.36 vs.\ 26.36) and R@L by 2.04 points (58.55 vs.\ 56.51). Remarkably, SignDPO also outperforms gloss-based methods on some metrics of this split, demonstrating that preference-based alignment can compensate for the absence of gloss supervision. On the development set, SignDPO achieves the highest B@4 of \textbf{28.95} and R@L of \textbf{58.62} among all reported gloss-free methods and is compatible with outstanding gloss-based methods.
 
\textbf{How2Sign.} Table~\ref{tab:how2sign_result} reports results on this challenging ASL benchmark. SignDPO sets a new state of the art across all metrics, achieving B@1 of \textbf{39.93}, B@4 of \textbf{15.58}, and R@L of \textbf{38.14}. These represent improvements of more than 10 B@1 points and 5 B@4 points over the previous best-performing method, FLa-LLM \citep{chen2024factorized}. The magnitude of these gains on a diverse, in-the-wild dataset underscores the benefit of multi-level preference signals for generalisation beyond constrained signing environments.
 
\textbf{OpenASL.}Table~\ref{tab:openasl_result} shows that SignDPO achieves \textbf{51.44} B@1, \textbf{24.35} B@4, and \textbf{46.29} R@L on the test set, and \textbf{52.83} B@1, \textbf{26.98} B@4, and \textbf{46.70} R@L on the development set. SignDPO surpasses Uni-Sign \citep{Li2025sign}, the previous state of the art, by 2.09 B@1, 1.21 B@4, and 3.07 R@L on the test set, and by 1.99 B@1, 2.82 B@4, and 2.12 R@L on the development set.

SignDPO consistently yields superior performance over existing baselines, validating the effectiveness of structured multi-level perturbations in aligning SLT models.

\begin{table}[htbp]
\centering
\caption{SLT results on CSL-Daily dataset. Best results are bolded, and \underline{second best} are underlined.}
\label{tab:csl_result}
\resizebox{\columnwidth}{!}{%
\begin{tabular}{clcccccc} 
\toprule
& \multirow{2}{*}{Method} & \multicolumn{3}{c}{DEV} & \multicolumn{3}{c}{TEST} \\
\cmidrule(lr){3-5}\cmidrule(lr){6-8}
& & B@1 & B@4 & R@L & B@1 & B@4 & R@L \\
\midrule
\multirow{6}{*}{\rotatebox{90}{Gloss-based}}
& SLRT \citep{camgoz2020sign}       & 37.47 & 11.88 & 37.96 & 37.38 & 11.79 & 36.74 \\
& SignBT \citep{zhou2021improving}   & 51.46 & 20.80 & 49.49 & 51.42 & 21.34 & 49.31 \\
& MMTLB \citep{chen2022simple}      & 53.81 & 24.42 & 53.38 & 53.31 & 23.92 & 53.25 \\
& SLTUNET \citep{zhang2023sltunet}  & -- & 23.99 & 53.58 & 54.98 & 25.01 & 54.08 \\
& TS-SLT \citep{chen2022two}        & \underline{55.21} & \underline{25.76} & \underline{55.10} & \underline{55.44} & \underline{25.79} & \underline{55.72} \\
& CV-SLT \citep{zhao2024conditional} & \textbf{58.05} & \textbf{28.24} & \textbf{56.36} & \textbf{58.29} & \textbf{28.94} & \textbf{57.06} \\
\midrule
\multirow{11}{*}{\rotatebox{90}{Gloss-free}}
& SLRT \citep{camgoz2020sign}       & 21.03 & 4.04 & 20.51 & 20.00 & 3.03 & 19.67 \\
& GASLT \citep{yin2023gloss}        & -- & -- & -- & 19.90 & 4.07 & 20.35 \\
& MSLU \citep{zhou2024scaling}       & 33.28 & 10.27 & 33.13 & 33.97 & 11.42 & 33.80 \\
& NSLT \citep{camgoz2018neural}      & 34.22 & 7.96 & 34.28 & 34.16 & 7.56 & 34.54 \\
& GFSLT-VLP \citep{zhou2023gloss}   & 39.20 & 11.07 & 36.70 & 39.37 & 11.00 & 36.44 \\
& FLa-LLM \citep{chen2024factorized} & -- & -- & -- & 37.13 & 14.20 & 37.25 \\
& $C^{2}$RL \citep{chen2025c}       & -- & -- & -- & 49.32 & 21.61 & 48.21 \\
& Uni-Sign \citep{Li2025sign}       & \underline{55.30} & \underline{26.25} & \underline{56.03} & \underline{55.08} & \underline{26.36} & \underline{56.51} \\
& SignLLM \citep{gong2024llms}      & 42.45 & 12.23 & 39.18 & 39.55 & 15.75 & 39.91 \\
& Sign2GPT \citep{wong2024sign2gpt}  & -- & -- & -- & 41.75 & 15.40 & 42.36 \\
\cmidrule{1-8}
& \textbf{SignDPO} & \textbf{57.87} & \textbf{28.95} & \textbf{58.62} & \textbf{57.43} & \textbf{28.36} & \textbf{58.55} \\
\bottomrule
\end{tabular}%
}
\end{table}

\begin{table}[htbp]
\centering
\caption{SLT results on How2Sign dataset. Best results are bolded, and \underline{second best} are underlined.}
\label{tab:how2sign_result}
\resizebox{\columnwidth}{!}{%
\begin{tabular}{lccccc}
\toprule
\multirow{2}{*}{Method} & \multicolumn{5}{c}{TEST} \\
\cmidrule(lr){2-6}
& B@1 & B@2 & B@3 & B@4 & R@L \\
\midrule
GloFE-VN \citep{lin2023gloss} & 14.90 & 7.30 & 3.90 & 2.20 & 12.60 \\
YouTube-ASL \citep{uthus2023youtube} & 37.80 & 24.10 & 16.90 & 12.40 & -- \\
MSLU \citep{zhou2024scaling} & 20.10 & 7.70 & -- & 2.40 & 17.20 \\
$C^{2}$RL \citep{chen2025c} & 29.10 & 18.60 & 12.90 & 9.40 & 27.00 \\
FLa-LLM \citep{chen2024factorized} & \underline{29.80} & \underline{19.00} & \underline{13.30} & \underline{9.70} & \underline{27.80} \\
\midrule
\textbf{SignDPO} & \textbf{39.93} & \textbf{27.44} & \textbf{21.07} & \textbf{15.58} & \textbf{38.14} \\
\bottomrule
\end{tabular}%
}
\end{table}

\begin{table}[htbp]
\centering
\captionsetup{font=footnotesize}
\caption{SLT results on OpenASL dataset. Best results are bolded, and \underline{second best} are underlined.}
\label{tab:openasl_result}
\resizebox{\columnwidth}{!}{%
\begin{tabular}{lcccccc}
\toprule
\multirow{2}{*}{Method} & \multicolumn{3}{c}{DEV} & \multicolumn{3}{c}{TEST} \\
\cmidrule(r){2-4} \cmidrule(l){5-7}
& B@1 & B@4 & R@L & B@1 & B@4 & R@L \\
\midrule
GloFE-VN \citep{lin2023gloss} & 21.06 & 6.68 & 21.37 & 21.56 & 7.06 & 21.75 \\
Conv-GRU \citep{camgoz2018neural} & 16.72 & 4.82 & 16.25 & 16.11 & 4.58 & 16.10 \\
I3D-transformer \citep{shi2022open} & 18.26 & 5.60 & 18.88 & 18.31 & 5.56 & 18.64 \\
OpenASL \citep{shi2022open} & 20.10 & 6.57 & 20.43 & 20.92 & 6.72 & 21.02 \\
Uni-Sign \citep{Li2025sign} & \underline{50.84} & \underline{24.16} & \underline{44.58} & \underline{49.35} & \underline{23.14} & \underline{43.22} \\
$C^{2}$RL \citep{chen2025c} & -- & -- & -- & 31.46 & 13.21 & 31.36 \\
\midrule
\textbf{SignDPO} & \textbf{52.83} & \textbf{26.98} & \textbf{46.70} & \textbf{51.44} & \textbf{24.35} & \textbf{46.29} \\
\bottomrule
\end{tabular}%
}
\end{table}

\subsection{Ablation study}
\begin{figure}[thbp!]
  \centering
  \includegraphics[scale=0.45]{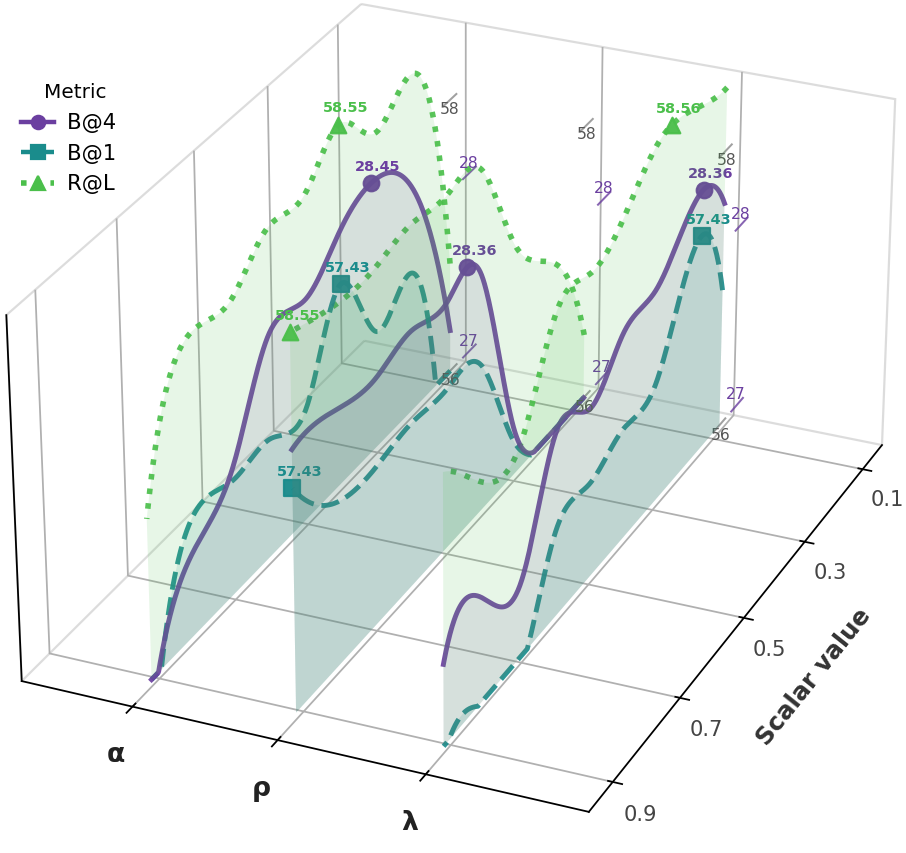}
    \caption{Hyperparameter sensitivity on the CSL-Daily test set of the spatial level weight $\alpha$, temporal level weight $\rho$, and language level weight $\lambda$ in the training objective (Eq. \ref{eq:joint_loss}). For each plane, the corresponding weight is varied over $\{0.1, \ldots, 0.9\}$. B@4 (purple, solid) uses its own scale, which is bounded by 30; B@1 (teal, dashed) and R@L (green, dotted) share the scale, which is bounded by 60. Peak performance values are annotated on each curve.}
  \label{fig:sensitivity}
\end{figure}

We conduct ablation studies to verify each design decision on CSL-Daily (test set). Results are summarised in Table \ref{tab:ablation} and the hyperparameter sensitivity analysis in Figure \ref{fig:sensitivity}.
 
\paragraph{Baseline comparisons (a).} We evaluate SignDPO against two foundational baselines in Table \ref{tab:ablation} (a): standard supervised fine-tuning (SFT) via MLE and vanilla DPO using only output-level preference pairs (language level DPO). Both are applied to the pre-trained model with the same training configuration. While SFT establishes the performance floor, reflecting the inherent limitations of cross-entropy training in capturing complex sign semantics with limited training epochs (10), vanilla DPO yields an improvement. This confirms that preference-based alignment provides a critical corrective signal that MLE-based supervision lacks. SignDPO achieves superior performance across all metrics. This performance gap underscores a fundamental "blindness" in output-only DPO: it treats mistranslations as purely linguistic errors while ignoring the perceptual misalignments in the visual input. By extending the preference signal into the spatial and temporal domains, SignDPO forces the model to align its internal visual-to-text mapping, effectively resolving perceptual failure modes that remain inaccessible to text-level supervision alone.
 
\paragraph{Multi-level loss ablation (b).} We ablate each component of the joint objective to assess its individual contribution to the final translation quality. As shown in Table \ref{tab:ablation} (b), removing the language modelling regularisation term ($\mathcal{L}_{\text{LM}}$) yields a light performance drop, confirming its role as a stabilising objective. Removing either the spatial ($\mathcal{L}_s$) or temporal ($\mathcal{L}_t$) loss leads to a measurable degradation, validating that explicit awareness of gestural layout and directional semantics is a primary driver of accuracy. Interestingly, while removing the language-level loss ($\mathcal{L}_l$) results in a smaller performance dip, relying on it in isolation yields the lowest scores among all single-component variants, suggesting that linguistic perturbations serve primarily as a complement to the more fundamental visual preference signals. Crucially, any single loss component used in isolation consistently underperforms the full joint objective, validating our multi-level design and proving that a holistic preference signal spanning spatial, temporal, and linguistic dimensions is essential for robust sign language translation.
 
\paragraph{Granularity ablation (c).} We examine the respective contributions of global and local granularities within the spatial and temporal preference components. 

When either the global or local granularity is ablated, the corresponding multi-level loss simplifies from a combo objective (the spatial level loss Eq. \ref{eq:sp_loss} and the temporal level Eq. \ref{eq:tmp_loss}) to a single-stream formulation :
\begin{equation}
\mathcal{L}_{\text{DPO}}^{\text{abla.}} = -\mathbb{E}_{(X,Y,X^-_{i})} \left[ \log \sigma \left( \beta \log \frac{\pi_\theta(Y|X)}{\pi_{\text{ref}}(Y|X)} - \beta \log \frac{\pi_\theta(Y|X^-_{i})}{\pi_{\text{ref}}(Y|X^-_{i})} \right) \right]
\end{equation}
where $i \in \{g, l\}$ represents the remaining branch. In these settings, the non-preferred log-ratio is computed exclusively using the specific perturbation strategy of the retained branch, while all other hierarchical components remain unchanged.

For the spatial loss, while removing the global branch ($\mathcal{L}_s$ w/o global) leads to a performance dip, omitting the local branch ($\mathcal{L}_s$ w/o local) triggers a much sharper decline. This contrast substantiates that attention-guided local perturbations provide a more potent preference signal than uniform global ones. A similar, yet more pronounced trend is observed in the temporal domain: removing the local temporal branch ($\mathcal{L}_t$ w/o local) results in the most severe single-component degradation in the entire ablation study. These results underscore that targeted temporal disruptions, applied precisely within the high-attention windows, are the primary mechanism for teaching the model to respect fine-grained gestural dynamics.

\begin{table}[htbp]
\centering
\caption{Ablation study on CSL-Daily (test set). (a) baseline comparisons against supervised fine-tuning (SFT) and vanilla DPO; (b) multi-level loss components and their combinations; (c) global vs.\ local granularity within spatial and temporal losses; (d) individual spatial perturbation strategies; (e) individual temporal perturbation strategies. The \colorbox{grayg}{worst} results and \colorbox{lightg}{second worst} results are highlighted. }
\label{tab:ablation}
\renewcommand{\arraystretch}{1.12} 
\resizebox{0.85\columnwidth}{!}{%
\begin{tabular}{l @{\hspace{3pt}}lccc} 
\toprule
& \textbf{Method} & \textbf{B@1} & \textbf{B@4} & \textbf{R@L} \\ \midrule

\small \multirow{3}{*}{\rotatebox{90}{\begin{tabular}{c}(a) \\ Base\end{tabular}}}
& SFT & \cellcolor{grayg}51.41 & \cellcolor{grayg}21.81 & \cellcolor{grayg}51.52 \\
& DPO & \cellcolor{lightg}54.20 & \cellcolor{lightg}25.10 & \cellcolor{lightg}55.12 \\
& SignDPO (Ours) & \textbf{57.43} & \textbf{28.36} & \textbf{58.55} \\ \midrule

\small \multirow{7}{*}{\rotatebox{90}{\begin{tabular}{c}(b) \\ Multi-level Loss\end{tabular}}}
& SignDPO w/o $\mathcal{L}_{\text{LM}}$ & 56.82 & 28.20 & 57.93 \\
& w/o $\mathcal{L}_s$ & 55.57 & 27.63 & \cellcolor{lightg}57.60 \\
& w/o $\mathcal{L}_t$ & \cellcolor{lightg}55.26 & \cellcolor{lightg}27.42 & 57.67 \\
& w/o $\mathcal{L}_l$ & 55.31 & 27.95 & 57.90 \\
& w $\mathcal{L}_s$ only & 56.38 & 27.93 & 57.95 \\ 
& w $\mathcal{L}_t$ only & 56.42 & 27.85 & 57.87 \\
& w $\mathcal{L}_l$ only & \cellcolor{grayg}54.20 & \cellcolor{grayg}25.10 & \cellcolor{grayg}55.12 \\ \midrule
 
\small \multirow{5}{*}{\rotatebox{90}{\begin{tabular}{c}(c) \\ Gran.\end{tabular}}}
& $\mathcal{L}_s$ w/o global & \cellcolor{lightg}54.98 & 27.77 & 57.82 \\
& $\mathcal{L}_s$ w/o local  & 55.16 & \cellcolor{lightg}26.43 & 57.58 \\
& $\mathcal{L}_t$ w/o global & 56.67 & 27.53 & \cellcolor{lightg}56.36 \\
& $\mathcal{L}_t$ w/o local  & \cellcolor{grayg}54.02 & \cellcolor{grayg}26.39 & \cellcolor{grayg}56.03 \\ \midrule

\small \multirow{5}{*}{\rotatebox{90}{\begin{tabular}{c}(c)\\ Spatial. \end{tabular}}}
& w/o Randomness & \cellcolor{grayg}56.58 & \cellcolor{grayg}27.97 & \cellcolor{grayg}58.03 \\
& w/o Blackness  & \cellcolor{lightg}56.70 & 28.05 & 58.27 \\
& w/o Sparse mask & 56.73 & \cellcolor{lightg}28.00 & 58.15 \\
& w/o ROI mask  & 56.71 & 28.05 & \cellcolor{lightg}58.13 \\ \midrule

\small \multirow{2}{*}{\rotatebox{90}{\begin{tabular}{c}(e)\\ Temp.\end{tabular}}}
& w/o Shuffle & \cellcolor{lightg}56.88 & \cellcolor{lightg}28.02 & \cellcolor{lightg}58.20 \\
& w/o Reverse  & \cellcolor{grayg}56.40 & \cellcolor{grayg}27.64 & \cellcolor{grayg}57.87 \\ \bottomrule
\end{tabular}%
}
\label{tab:ablation}
\end{table}
 
\paragraph{Spatial strategy ablation (d).} We evaluate the individual contributions of our four spatial perturbation strategies by removing each while retaining the others. Each perturbation strategy is selected with equal probability of 25\%, and if the selected strategy is the ablation target, the original unperturbed sample is used. While all four strategies contribute positively to the final result, their impact varies in magnitude. Notably, removing the randomness strategy triggers the most significant performance decline. This result attests to the high contrastive value of exposing the model to fully mismatched spatial configurations, which effectively penalises the misidentification of skeleton-to-text mappings. In contrast, the removal of the ROI mask or sparse mask results in more modest degradations; however, their inclusion consistently elevates scores, particularly ROUGE-L in the case of ROI, validating a comprehensive approach to spatial integrity. Given their negligible computational overhead, the full suite of strategies is retained to ensure maximum robustness against spatial noise.
 
\paragraph{Temporal strategy ablation (e).} We compare the impact of our two temporal perturbation techniques. Each perturbation strategy is selected with equal probability of 50\%, and if the selected strategy is the ablation target, the original unperturbed sample is used. Results show that removing the reverse strategy leads to a more pronounced performance drop compared to omitting shuffle. This disparity underscores that training the model to distinguish the correct directionality of gestures, the "temporal flow", is more semantically valuable than simply identifying chronological order. 

\paragraph{Hyperparameter sensitivity.} We investigate the sensitivity of \\ SignDPO to the weighting of the spatial ($\alpha$), temporal ($\rho$), and linguistic ($\lambda$) preference signals across the range $[0.1, 0.9]$. As illustrated in Fig. \ref{fig:sensitivity}, the model demonstrates robustness to moderate perturbations around the optima, with metrics exhibiting a characteristic "bell-shaped" response. Spatial and temporal level weights ($\alpha, \rho$): These components require more precise calibration to balance the visual preference signals. Performance for $\alpha$ peaks early at around 0.4, while $\rho$ achieves its optimum at a higher value of 0.9. The sharper declines observed when these weights exceed their optimal points suggest that disproportionately large visual perturbations can overwhelm the primary translation objective. Language level weight ($\lambda$): In contrast, $\lambda$ displays a relatively stable response across the middle range, peaking at 0.2. This validates its role as a consistent regulariser that effectively aligns the output distribution with human-preferred linguistic patterns without the volatility associated with visual-level perturbations.

\section{Conclusion}
In this work, we introduced SignDPO, a novel multi-level preference optimisation framework designed to resolve the perceptual and linguistic failure modes inherent in standard SLT models. By moving beyond the "output-only" limitation of traditional DPO, our approach effectively aligns the model's internal visual-to-text mapping through structured perturbations across spatial, temporal, and linguistic dimensions. Our experimental results on three major benchmarks, CSL-Daily, How2Sign, and OpenASL, consistently demonstrate that SignDPO sets a new performance ceiling for gloss-free SLT. Ablation studies underscore that targeted local perturbations, guided by cross-attention saliency, provide a significantly more potent training signal than global distortions. Furthermore, our results validate that while linguistic alignment is a valuable regulariser, the core of robust SLT lies in the model's sensitivity to fine-grained gestural dynamics. We believe the paradigm of input-level preference learning presented in SignDPO offers a principled path forward for other multimodal tasks where the integrity of the input modality is directly tied to the semantic faithfulness of the generation.

\appendix

\section{Language Perturbation Model}
To provide a scalable and high-quality source of linguistic preference signals without manual annotation, we develop a dedicated language perturbation model based on GPT-4.1-nano. This model is specifically trained to simulate the failure modes of SLT systems by generating semantically adjacent but inaccurate translations.

\subsection{Quality Scoring Metrics}
We construct the training data $\mathcal{D}$ for the perturbation model by scoring the raw outputs
($\mathcal{T}_{\text{raw}}$) of our SLT policy model on the dev and test sets of CSL-Daily, How2Sign, and OpenASL. Each sample is quantified across three dimensions. While the weights are consistent across datasets, the underlying calculation of sub-metrics varies to accommodate the linguistic characteristics of English and Chinese.

\paragraph{Dimension 1: Adequacy ($A$)}
Adequacy measures the retention of source semantics, primarily focusing on
recall-based metrics to identify information omission:
\begin{equation}
A = 0.4 \cdot \mathrm{BS}_R + 0.3 \cdot \mathrm{Sim}_{\mathrm{char}}
  + 0.2 \cdot \mathrm{Cov}_{\mathrm{kw}} + 0.1 \cdot \mathrm{Len}_A
\end{equation}

\noindent\textbf{$\mathrm{BS}_R$ --- BERTScore Recall.}
Computed using a pre-trained contextual language model (\texttt{bert-base-chinese} for
Chinese; \\ \texttt{roberta-large} for English), $\mathrm{BS}_R$ measures the degree to
which every token in the reference is semantically covered by the prediction.
Each reference token $r_j$ is matched to its most similar prediction token via cosine
similarity in embedding space, and the mean of these maximum similarities is rescaled
against a language-specific baseline to yield a score in $[0,1]$.
A high $\mathrm{BS}_R$ indicates that key reference content has not been omitted.

\noindent\textbf{$\mathrm{Sim}_{\mathrm{char}}$ --- Character-level Jaccard Similarity.}
Let $C_p$ and $C_r$ denote the sets of unique characters (excluding spaces)
in the prediction and reference respectively. The score is
\begin{equation}
\mathrm{Sim}_{\mathrm{char}} = \frac{|C_p \cap C_r|}{|C_p \cup C_r|}
\end{equation}
This set-based measure is language-agnostic and captures the proportion of
character types shared between the two sentences, providing a lightweight
lexical overlap signal robust to word-order variation.

\noindent\textbf{$\mathrm{Cov}_{\mathrm{kw}}$ --- Keyword Coverage.}
A set of up to four content keywords $\mathcal{K}_r$ is extracted from the reference,
and coverage is defined as
\begin{equation}
\mathrm{Cov}_{\mathrm{kw}} = \frac{|\{k \in \mathcal{K}_r : k \in \text{pred}\}|}{|\mathcal{K}_r|}
\end{equation}
For \emph{Chinese} (CSL-Daily), \texttt{jieba} POS tagging retains nouns
(\texttt{n}, \texttt{nr}, \texttt{ns}, \texttt{nt}, \texttt{nz}), verbs (\texttt{v},
\texttt{vn}), adjectives (\texttt{a}), numerals (\texttt{m}), and time words
(\texttt{t}) as candidates; adjacent content-word bigrams (e.g.\ numeral\,+\,noun)
are additionally considered.
For \emph{English} (How2Sign, OpenASL), NLTK POS tagging retains tokens whose tags
begin with \texttt{NN}, \texttt{VB}, \texttt{JJ}, \texttt{RB}, or \texttt{CD} and
are not stopwords; adjacent adjective--noun and noun--noun pairs are also considered.
In both cases, candidates are ranked by a composite score reflecting POS category,
token length, within-sentence frequency, and position, and the top-4 non-overlapping
tokens are selected.

\noindent\textbf{$\mathrm{Len}_A$ --- Length Adequacy.}
\begin{equation}
\mathrm{Len}_A = \min\!\left(\frac{\mathrm{len}(\mathrm{pred})}{\mathrm{len}(\mathrm{ref})},\;1.0\right)
\end{equation}
Length is measured in characters for Chinese and whitespace-delimited tokens for
English. The clamping at $1.0$ ensures over-long predictions are not penalised
here (that is handled by $\mathrm{Len}_F$), while shorter predictions are penalised
proportionally, reflecting potential content omission.

\paragraph{Dimension 2: Faithfulness ($F$)}
Faithfulness penalises hallucinations and irrelevant additions by focusing on
precision-based metrics:
\begin{equation}
F = 0.4 \cdot \mathrm{BS}_P + 0.2 \cdot \mathrm{Len}_F
  + 0.2 \cdot \mathrm{Cons}_{\mathrm{num}} + 0.2 \cdot \mathrm{Cons}_{\mathrm{ent}}
\end{equation}

\noindent\textbf{$\mathrm{BS}_P$ --- BERTScore Precision.}
The precision counterpart of $\mathrm{BS}_R$: each prediction token $p_i$ is matched
to its closest reference token in embedding space, and the mean maximum similarity is
rescaled with the same language-specific baseline. A low $\mathrm{BS}_P$ indicates
that prediction tokens are semantically distant from any reference token, i.e.\ the
model has generated content not supported by the reference.

\noindent\textbf{$\mathrm{Len}_F$ --- Length Faithfulness.}
\begin{equation}
\mathrm{Len}_F =
\begin{cases}
\dfrac{\mathrm{len}(\mathrm{ref})}{\mathrm{len}(\mathrm{pred})} &
  \text{if } \dfrac{\mathrm{len}(\mathrm{pred})}{\mathrm{len}(\mathrm{ref})} > 1.2 \\[6pt]
1.0 & \text{otherwise}
\end{cases}
\end{equation}
No penalty is applied when prediction length is within $1.2\times$ the reference;
beyond this threshold the score decreases inversely with over-generation, discouraging
verbosity and repetition.

\noindent\textbf{$\mathrm{Cons}_{\mathrm{num}}$ --- Numerical Consistency.}
Let $\mathcal{N}_r$ and $\mathcal{N}_p$ be the sets of digit sequences extracted
from the reference and prediction respectively. Then
\begin{equation}
\mathrm{Cons}_{\mathrm{num}} =
\begin{cases}
\dfrac{|\mathcal{N}_r \cap \mathcal{N}_p|}{|\mathcal{N}_r|} & \text{if } |\mathcal{N}_r| > 0 \\[6pt]
1.0 & \text{if } |\mathcal{N}_r| = 0 \text{ and } |\mathcal{N}_p| = 0 \\[6pt]
0.8 & \text{if } |\mathcal{N}_r| = 0 \text{ and } |\mathcal{N}_p| > 0
\end{cases}
\end{equation}
The partial penalty of $0.8$ in the third case discourages hallucinating numeric
content absent from the reference.

\noindent\textbf{$\mathrm{Cons}_{\mathrm{ent}}$ --- Named-Entity Consistency.}
For \emph{English}, NLTK's \texttt{ne\_chunk} extracts named-entity spans (persons,
locations, organisations); for \emph{Chinese}, \texttt{jieba} POS labels \texttt{nr}
(person), \texttt{ns} (place), and \texttt{nt} (organisation) are used. Let
$\mathcal{E}_r$ and $\mathcal{E}_p$ denote the resulting entity sets. The score
follows the same set-ratio formula as $\mathrm{Cons}_{\mathrm{num}}$, penalising
missing reference entities and applying a mild $0.8$ penalty when the prediction
introduces entities absent from the reference.

\paragraph{Dimension 3: Fluency ($U$)}
Fluency assesses structural integrity and surface naturalness of the translation:
\begin{equation}
\begin{split}
U = \; & 0.25 \cdot \mathrm{O}_{\mathrm{BLEU}} + 0.25 \cdot R_L
        + 0.15 \cdot \mathrm{METEOR} \\
      & + 0.15 \cdot \mathrm{Sim}_{\mathrm{char}} + 0.1 \cdot \mathrm{Str}
        + 0.1 \cdot \mathrm{COMET}
\end{split}
\end{equation}
All sub-scores are normalised to $[0,1]$ before aggregation.

\noindent\textbf{$\mathrm{O}_{\mathrm{BLEU}}$ --- Overall BLEU.}
SacreBLEU with the \texttt{13a} tokeniser computes corpus-level $n$-gram precision
scores $\mathrm{BLEU}_n$ for $n \in \{1,2,3,4\}$. The overall score is their
geometric mean:
\begin{equation}
\mathrm{O}_{\mathrm{BLEU}} = \Bigl(\prod_{n=1}^{4} \mathrm{BLEU}_n\Bigr)^{1/4}
\end{equation}
$\mathrm{BLEU}_1$ captures unigram lexical accuracy while higher-order $n$-grams
reflect local grammatical fluency. A set-based fallback is used when SacreBLEU is
unavailable.

\noindent\textbf{$R_L$ --- ROUGE-L F-score.}
ROUGE-L measures the longest common subsequence (LCS) between prediction and
reference at the token level. The F-score ($\beta=1$) balances LCS-based precision
$P_{\mathrm{lcs}} = \mathrm{LCS}/|\text{pred}|$ and recall
$R_{\mathrm{lcs}} = \mathrm{LCS}/|\text{ref}|$. Unlike $n$-gram metrics, ROUGE-L
does not require contiguous matches, making it robust to paraphrases and word-order
variation. Scores are on a $[0,100]$ scale.

\noindent\textbf{METEOR.}
For \emph{English}, NLTK's METEOR aligns prediction and reference tokens via exact
match, stemming, and WordNet synonym lookup, making it sensitive to valid paraphrases
that BLEU would penalise. For \emph{Chinese}, METEOR is excluded due to the absence
of a reliable Chinese WordNet interface, and its weight is redistributed to the
remaining sub-metrics.

\noindent\textbf{$\mathrm{Sim}_{\mathrm{char}}$ --- Character-level Jaccard Similarity.}
The same set-based character overlap described under Adequacy is reused here as a
lightweight surface-form signal, capturing shared vocabulary at the character level
without requiring tokenisation.

\noindent\textbf{$\mathrm{Str}$ --- Structural Completeness.}
$\mathrm{Str}$ is a rule-based score averaged over two components. The
\emph{terminal punctuation} component assigns $1.0$ if the prediction ends with a
sentence-final mark (\texttt{.~!~?} for English; \texttt{%
\begin{CJK*}{UTF8}{gbsn}。！？\end{CJK*}} for Chinese),
$0.7$ for a clause-final mark (\texttt{,~;} for English;
\texttt{\begin{CJK*}{UTF8}{gbsn}，；\end{CJK*}} for Chinese), and $0.5$ otherwise.
The \emph{bracket balance} component assigns $1.0$ when opening brackets
(\texttt{(~[}~\begin{CJK*}{UTF8}{gbsn}（【\end{CJK*})
equal closing brackets
(\texttt{)~]}~\begin{CJK*}{UTF8}{gbsn}）】\end{CJK*}),
and $0.8$ otherwise. Together these heuristics penalise truncated or malformed
outputs without requiring a language model.

\noindent\textbf{COMET.}
Where available, the \texttt{wmt22-comet-da} model produces a reference-based quality
estimate grounded in human direct-assessment judgements. Its output is clamped to
$[0,1]$. When COMET cannot be loaded (e.g.\ during large-scale perturbation
generation), this term defaults to $0$ and its $0.1$ weight is absorbed by the
remaining sub-metrics through the normalisation step.

\subsection{SFT Prompt Design}
The perturbation model is fine-tuned using Supervised Fine-Tuning (SFT) with a
specific prompt template designed to condition text generation on the quality scores.

\paragraph{System Prompt:}
\begin{quote}
\itshape
You are an assistant that generates text based on scoring conditions. Given a
reference text (ref\_text) and three scores: \\
- adequacy\_final: 0$\sim$1, higher means better semantic adequacy and key
information retention; \\
- faithfulness\_final: 0$\sim$1, higher means more faithful output with no
irrelevant additions; \\
- fluency\_final: 0$\sim$1, higher means better fluency and structural completeness. \\
Based on these three scores, simulate the target output (pre\_text) in terms of
language style and degree of information retention. Note: Do not repeat the scores
themselves --- only output the generated result.
\end{quote}

\paragraph{Training Sample Format.}
Training examples are formatted as JSONL message triples:
\begin{itemize}
  \item \textbf{User:} \texttt{ref\_text: [Reference] adequacy\_final: [A]
        faithfulness\_final: [F] fluency\_final: [U]}
  \item \textbf{Assistant:} \texttt{[Perturbed translation]} $Y^-$
\end{itemize}

\subsection{Training and Inference Configuration}
We fine-tune the GPT-4.1-nano model for 3 epochs with a batch size of 8 and a 10\%
validation split on the scored dataset. During the preference alignment phase of
SignDPO, we generate $Y^-_l$ by sampling $(A, F, U)$ triples from the empirical
distribution of the training set. This ensures that the generated non-preferred
samples accurately reflect the actual distribution of errors the policy model is
likely to encounter.

\section{Dataset Statistics}
Table~\ref{tab:datasets} details the sign language benchmarks utilised in this
research, categorised by language, linguistic granularity, sample count, and storage
requirements for RGB and skeletal data. While American Sign Language is represented
by YouTube-ASL, WLASL, How2Sign, and OpenASL, Chinese Sign Language is covered by
CSL-News and CSL-Daily. Skeletal data offers a tenfold reduction in storage compared
to RGB video, translating into diminished computational overhead and accelerated data
throughput. The efficiency gains in file size and latency are further quantified in
Table~\ref{tab:modality_comparison}.

\begin{table}[htbp]
\centering
\caption{Dataset statistics (sizes in GB).}
\label{tab:datasets}
\setlength{\tabcolsep}{4pt}
\resizebox{\columnwidth}{!}{%
\begin{tabular}{llccccc}
\toprule
 & Dataset & Language & Level & \# Samples & Size (RGB) & Size (Skeleton) \\
\midrule
\multirow{2}{*}{Pre-train}
& YouTube-ASL \cite{uthus2023youtube}  & American & Sentence & 530,161 & 2,320.87 & 108.30 \\
& CSL-News \cite{Li2025sign}           & Chinese  & Sentence & 722,715 & 5,470.86 & 255.29 \\
\midrule
\multirow{3}{*}{Fine-tune}
& How2Sign \cite{duarte2021how2sign}   & American & Sentence & 35,263  & 329.00   & 15.58  \\
& OpenASL \cite{shi2022open}           & American & Sentence & 98,419  & 638.03   & 29.78  \\
& CSL-Daily \cite{zhou2021improving}   & Chinese  & Sentence & 20,654  & 92.80    & 4.27   \\
\bottomrule
\end{tabular}%
}
\end{table}

\begin{table}[htbp]
\centering
\caption{Comparison of RGB and skeletal data.}
\label{tab:modality_comparison}
\resizebox{\columnwidth}{!}{%
\begin{tabular}{ccc}
\toprule
Modality & Avg.\ size per sample (KB) & Loading time per sample (ms) \\
\midrule
RGB      & 4714.84 & 455.35 \\
Skeletal & 437.18  & 9.58   \\
\bottomrule
\end{tabular}%
}
\end{table}

\section{Qualitative Comparison}
Tables \ref{tab:qualitative_csl}, \ref{tab:qualitative_how2sign}, and \ref{tab:qualitative_openasl} illustrate the contrast between reference translations and the outputs of different optimisation strategies.

\begin{CJK}{UTF8}{gbsn}
\begin{table}[htbp]
\centering\small
\caption{Qualitative comparison on \textbf{CSL-Daily}. Text highlighted in
  \smash{\colorbox{hlgreen}{green}} denotes exact matches,
  \smash{\colorbox{hlyellow}{yellow}} semantically similar expressions,
  \smash{\colorbox{hlpurple}{purple}} structural omissions, and
  \smash{\colorbox{hlred}{red}} incorrect expressions.}
\label{tab:qualitative_csl}
\renewcommand{\arraystretch}{1.4}
\begin{tabularx}{\columnwidth}{@{} l X @{}}
\toprule
\textbf{Source} & \textbf{Sentence (CSL-Daily)} \\
\midrule

\textbf{Reference}
  & 天气预报明天下雪,多穿衣服。 \\
  & \textit{(The weather forecast says it will snow tomorrow, wear more clothes.)} \\[3pt]

\textbf{SFT}
  & \hgreen{天气预报}\hred{说后天}\hgreen{下雪,}\hred{注意}\hgreen{穿衣服。} \\[1pt]
  & \textit{\hgreene{The weather forecast says} \hrede{the day after tomorrow} \hgreene{it will snow,} \hrede{pay attention to} \hgreene{wearing clothes.}} \\[3pt]

\textbf{DPO}
  & \hgreen{天气预报明天}\hyellow{会}\hgreen{下雪,}\hyellow{记得}\hgreen{多穿衣服。} \\[1pt]
  & \textit{\hgreene{The weather forecast says tomorrow} \hyellowe{it will probably} \hgreene{snow,} \hyellowe{remember to} \hgreene{wear more clothes.}} \\[3pt]

\textbf{SignDPO}
  & \hgreen{天气预报明天下雪,多穿衣服。} \\[1pt]
  & \textit{\hgreene{The weather forecast says it will snow tomorrow, wear more clothes.}} \\[5pt]

\midrule

\textbf{Reference}
  & 老师在多年的教学工作中取得了丰富的经验。 \\
  & \textit{(The teacher gained rich experience through many years of teaching work.)} \\[3pt]

\textbf{SFT}
  & \hgreen{老师}\hred{教了很多年},\hred{在工作中}\hyellow{获得了丰硕的}\hpurple{教学} \hgreen{经验}。 \\[1pt]
  & \textit{\hgreen{The teacher}\hred{taught for many years,} \hrede{in work} \hyellowe{obtained fruitful} \hpurplee{teaching} \hgreen{experience.}} \\[3pt]

\textbf{DPO}
  & \hgreen{老师在多年的}\hyellow{工作中}\hyellow{积累了}\hgreen{丰富的经验}。 \\[1pt]
  & \textit{\hgreene{The teacher,} \hyellowe{through many years of work,} \hyellowe{accumulated} \hgreene{rich experience.}} \\[3pt]

\textbf{SignDPO}
  & \hgreen{老师在}\hyellow{学习和}\hgreen{工作中取得了}\hyellow{丰硕}\hgreen{的经验}。 \\[1pt]
  & \textit{\hgreene{The teacher gained} \hyellowe{fruitful} \hgreene{experience through} \hyellowe{study and} \hgreene{work.}} \\[5pt]

\midrule

\textbf{Reference}
  & 上课时,老师总是用温和的目光看着大家,使我们感到亲切。 \\
  & \textit{(In class, the teacher always looked at everyone with gentle eyes, making us feel warm.)} \\[3pt]

\textbf{SFT}
  & \hgreen{上课时,老师}\hred{经常}\hgreen{用}\hyellow{温柔的眼神}\hgreen{看着}\hred{学生},\hred{让人} \hgreen{感到亲切}。 \\[1pt]
  & \textit{\hgreene{In class, the teacher} \hrede{often} \hgreene{looked at} \hrede{students} \hgreene{with} \hyellowe{warm eyes,} \hrede{making people} \hgreene{feel warm.}} \\[3pt]

\textbf{DPO}
  & \hgreen{上课时,老师总是}\hyellow{温柔地}\hgreen{看着大家,让我们感到亲切}。 \\[1pt]
  & \textit{\hgreene{In class, the teacher always} \hyellowe{kindly} \hgreene{looked at everyone, making us feel warm.}} \\[3pt]

\textbf{SignDPO}
  & \hgreen{上课时,老师总是}\hyellow{温和地}\hgreen{看着大家,使我们感到亲切}。 \\[1pt]
  & \textit{\hgreene{In class, the teacher always} \hyellowe{gently} \hgreene{looked at everyone, making us feel warm.}} \\

\bottomrule
\end{tabularx}
\end{table}
\end{CJK}

\begin{table}
\centering\small
\caption{Qualitative comparison on \textbf{How2Sign}. Color legend same as
  Table~\ref{tab:qualitative_csl}.}
\renewcommand{\arraystretch}{1.4}
\begin{tabularx}{\columnwidth}{@{} l X @{}}
\toprule
\textbf{Source} & \textbf{Sentence (How2Sign)} \\
\midrule

\textbf{Reference}
  & Today we're going to be learning how to play Portal, a game by Valve Software. \\[3pt]
\textbf{SFT}
  & \hgreene{Today} \hrede{I will show you} \hgreene{how to play Portal,}
    \hpurplee{a game by Valve Software.} \\[1pt]
  & \textit{Subject shifted to first person; ``a game by Valve Software'' structurally omitted.} \\[3pt]
\textbf{DPO}
  & \hgreene{Today} \hrede{I will show you} \hgreene{how to play Portal, a game}
    \hyellowe{made by} \hgreene{Valve Software.} \\[1pt]
  & \textit{Subject still shifted to first person; ``by'' paraphrased as ``made by''.} \\[3pt]
\textbf{SignDPO}
  & \hgreene{Today we're going to} \hyellowe{talk about} \hgreene{learning how to play Portal,
    a game} \hyellowe{from} \hgreene{Valve Software.} \\[5pt]

\midrule

\textbf{Reference}
  & So, we've got to find a way to get to the exit. \\[3pt]
\textbf{SFT}
  & \hgreene{So,} \hrede{we need} \hgreene{to find} \hrede{the exit somehow.} \\[1pt]
  & \textit{``A way to get to'' dropped; meaning partially preserved.} \\[3pt]
\textbf{DPO}
  & \hgreene{So,} \hrede{what we're going to do is we're going}
    \hgreene{to find a way to get to the exit.} \\[3pt]
\textbf{SignDPO}
  &  \hgreene{So, we've got to find} \hyellowe{a path} \hgreene{to get to the exit.}  \\[5pt]

\midrule

\textbf{Reference}
  & So, I would advise putting a lot of your practice time into the sand wedge. \\[3pt]
\textbf{SFT}
  & \hgreene{So,} \hrede{you should use the sand wedge more when you practice.} \\[1pt]
  & \textit{Subject shifted; ``advise'' and ``practice time'' emphasis lost; structure reordered.} \\[3pt]
\textbf{DPO}
  & \hgreene{I} \hyellowe{strongly recommend} \hgreene{that you}
    \hpurplee{practice} \hrede{this.}  \\[3pt]
\textbf{SignDPO}
  & \hgreene{So, I would} \hyellowe{recommend} \hgreene{putting} \hyellowe{more of}
    \hgreene{your practice time into the sand wedge.} \\

\bottomrule
\end{tabularx}
\label{tab:qualitative_how2sign}
\end{table}

Table \ref{tab:qualitative_csl} presents three CSL-Daily examples that illustrate the progressive improvement in translation quality across SFT, DPO, and SignDPO. In the first example, SFT introduces two factual errors, substituting the temporal reference ``tomorrow'' with ``the day after tomorrow'' and inserting the redundant modifier ``pay attention to'', while preserving the core lexical content. DPO corrects both errors but adds hedging modifiers (``probably'', ``remember to'') not present in the reference, reflecting residual uncertainty in the model's predictions. SignDPO produces an exact match, demonstrating that preference optimisation successfully suppresses both factual substitutions and unnecessary additions. The second example reveals a more severe SFT failure: the sentence structure is entirely restructured, the main verb ``gained'' is replaced with the incorrect ``taught'', and the compound modifier ``teaching work'' is split and partially omitted, yielding a translation that conveys a substantially different meaning. DPO recovers the correct subject--verb--object structure and retains the key phrase ``rich experience'', though it generalises ``teaching work'' to the broader ``work'' and substitutes ``accumulated'' for ``gained''. SignDPO likewise omits ``teaching'' but introduces ``study and'', a semantically plausible yet unsupported addition; the core meaning is nonetheless largely preserved. The third example demonstrates SFT's tendency toward imprecise scope: ``always'' is weakened to ``often'', the collective referent ``everyone'' is narrowed to ``students'', and the personal expression ``making us feel warm'' is depersonalised to ``making people feel warm''. DPO and SignDPO both recover the correct adverbial scope and referent, differing only in word choice (``kindly'' vs.\ ``gently'') — both acceptable paraphrases of the reference's ``with gentle eyes''. Taken together, these examples show that SFT is prone to factual substitution, structural distortion, and scope narrowing, while DPO largely resolves structural errors at the cost of minor lexical imprecision. SignDPO consistently achieves the closest alignment with the reference, confirming the effectiveness of multi-level preference optimisation for Chinese SLT.

The results for the How2Sign dataset, presented in Table \ref{tab:qualitative_how2sign}, further highlight the differences in how each optimization strategy handles complex instructional English. In the first example, both SFT and DPO struggle with the narrative perspective, shifting the inclusive "we're going to" to a more authoritative first-person "I will show you." SFT additionally suffers from structural omission, completely dropping the appositive phrase "a game by Valve Software." While DPO recovers this phrase, it introduces a minor lexical paraphrase ("made by"). In contrast, SignDPO correctly identifies the inclusive subject and retains the instructional "learning" component, yielding the most faithful reproduction of the reference's introductory tone. The second example showcases SFT's tendency toward information compression, where the phrase "a way to get to" is collapsed into a generic "somehow," losing the specific procedural intent of the source. DPO improves upon this by recovering the full syntactic structure but introduces significant verbosity ("what we're going to do is..."), which, while grammatically correct, deviates from the concise nature of the reference. SignDPO achieves a balance, preserving the original "got to find" urgency while using a semantically similar noun ("path" instead of "way") to maintain the core meaning without unnecessary wordiness. In the third example, a technical instruction regarding sports equipment, SFT fails to capture the advisory nature of the sentence, replacing the nuanced "I would advise" with a blunt "you should use." This results in the loss of the specific emphasis on "practice time." DPO exhibits a different failure mode here: it over-optimizes for a "standard" recommendation template ("I strongly recommend that you..."), which leads to a severe hallucination/omission where the specific object ("sand wedge") is replaced by a vague "this." SignDPO excels in this complex scenario, retaining the conditional "I would," the specific recommendation of "putting," and the technical "sand wedge" terminology, differing from the reference only by a minor additive modifier ("more of"). Overall, the How2Sign comparisons demonstrate that while SFT often truncates instructions and DPO frequently falls into "template-matching" hallucinations, SignDPO maintains the highest level of terminological density and perspectival accuracy. This confirms that multi-level preference optimization is particularly effective at preserving the specific instructional nuances required for high-quality English sign language translation.

\begin{table}[t]
\centering\small
\caption{Qualitative comparison on \textbf{OpenASL}. Color legend same as Table~\ref{tab:qualitative_csl}.}
\renewcommand{\arraystretch}{1.4}
\begin{tabularx}{\columnwidth}{@{} l X @{}}
\toprule
\textbf{Source} & \textbf{Sentence (OpenASL)} \\
\midrule
\textbf{Reference}
  & We encourage the committee to come together and advocate for mental health. \\[3pt]
\textbf{SFT}
  & \hgreene{We} \hrede{want} \hgreene{the committee to} \hrede{unite}
    \hgreene{for mental health.} \\[1pt]
  & \textit{``Encourage'' replaced with ``want''; ``come together and advocate'' collapsed to ``unite''.} \\[3pt]
\textbf{DPO}
  & \hrede{We're a part of a committee that is} \hyellowe{committed to}
    \hgreene{mental health.} \\[3pt]
\textbf{SignDPO}
  &   \hgreene{We encourage the committee to} \hyellowe{gather} \hgreene{and}
    \hyellowe{support} \hgreene{mental health.} \\[5pt]

\midrule
\textbf{Reference}
  & The business agreement would be in regards to the certification program. \\[3pt]
\textbf{SFT}
  & \hgreene{The business} \hyellowe{deal} \hrede{is about getting certified.} \\[1pt]
  & \textit{``Agreement'' paraphrased as ``deal''; ``certification program'' incorrectly compressed.} \\[3pt]
\textbf{DPO}
  & \hgreene{The business} \hrede{happens because it is} \hyellowe{about a}
    \hgreene{certification program.} 
\\[3pt]
\textbf{SignDPO}
  & \hgreene{The business} \hyellowe{deal} \hgreene{would be} \hyellowe{related to}
    \hgreene{the certification program.} \\[5pt]

\midrule
\textbf{Reference}
  & We will be happy to respond, give you support and listen to your concerns. \\[3pt]
\textbf{SFT}
  & \hgreene{We} \hrede{are glad to} \hgreene{respond} \hyellowe{and give support}
    \hgreene{and listen to} \hrede{what you say.} \\[1pt]
  & \textit{``Will be happy'' shifted to present tense; ``you'' dropped from support;
    ``concerns'' replaced with ``what you say''.} \\[3pt]
\textbf{DPO}
  & \hgreene{We will be} \hyellowe{pleased} \hgreene{to respond,} \hyellowe{provide}
    \hgreene{support and listen to your concerns.} \\[3pt]
\textbf{SignDPO}
  & \hyellowe{We are} \hyellowe{pleased} \hgreene{to respond, support, and listen to
    your concerns.} \\
\bottomrule
\end{tabularx}
\label{tab:qualitative_openasl}
\end{table}

\begin{table*}[htbp]
\centering
\caption{SLT results on CSL-Daily dataset. Best results are \textbf{bolded}, and \underline{second best} are underlined.}
\label{tab:csl_result}
\resizebox{0.7\textwidth}{!}{%
\begin{tabular}{clcccccccccc} 
\toprule
& \multirow{2}{*}{Method} & \multicolumn{5}{c}{DEV} & \multicolumn{5}{c}{TEST} \\
\cmidrule(lr){3-7}\cmidrule(lr){8-12}
& & B@1 & B@2 & B@3 & B@4 & R@L & B@1 & B@2 & B@3 & B@4 & R@L \\
\midrule
\multirow{6}{*}{\rotatebox{90}{Gloss-based}}
& SLRT \citep{camgoz2020sign}       & 37.47 & 24.67 & 16.86 & 11.88 & 37.96 & 37.38 & 24.36 & 16.55 & 11.79 & 36.74 \\
& SignBT \citep{zhou2021improving}   & 51.46 & 37.23 & 27.51 & 20.80 & 49.49 & 51.42 & 37.26 & 27.76 & 21.34 & 49.31 \\
& MMTLB \citep{chen2022simple}      & 53.81 & 40.84 & 31.29 & 24.42 & 53.38 & 53.31 & 40.41 & 30.87 & 23.92 & 53.25 \\
& SLTUNET \citep{zhang2023sltunet}  & --    & --    & --    & 23.99 & 53.58 & 54.98 & 41.44 & 31.84 & 25.01 & 54.08 \\
& TS-SLT \citep{chen2022two}        & \underline{55.21} & \underline{42.31} & \underline{32.71} & \underline{25.76} & \underline{55.10} & \underline{55.44} & \underline{42.59} & \underline{32.87} & \underline{25.79} & \underline{55.72} \\
& CV-SLT \citep{zhao2024conditional} & \textbf{58.05} & \textbf{44.73} & \textbf{35.14} & \textbf{28.24} & \textbf{56.36} & \textbf{58.29} & \textbf{45.15} & \textbf{35.77} & \textbf{28.94} & \textbf{57.06} \\
\midrule
\multirow{10}{*}{\rotatebox{90}{Gloss-free}}
& SLRT \citep{camgoz2020sign}       & 21.03 & 9.97  & 5.96  & 4.04  & 20.51 & 20.00 & 9.11  & 4.93  & 3.03  & 19.67 \\
& GASLT \citep{yin2023gloss}        & --    & --    & --    & --    & --    & 19.90 & 9.94  & 5.98  & 4.07  & 20.35 \\
& MSLU \citep{zhou2024scaling}       & 33.28 & 21.31 & --    & 10.27 & 33.13 & 33.97 & 22.20 & --    & 11.42 & 33.80 \\
& NSLT \citep{camgoz2018neural}      & 34.22 & 19.72 & 12.24 & 7.96  & 34.28 & 34.16 & 19.57 & 11.84 & 7.56  & 34.54 \\
& GFSLT-VLP \citep{zhou2023gloss}   & 39.20 & 25.02 & 16.35 & 11.07 & 36.70 & 39.37 & 24.93 & 16.26 & 11.00 & 36.44 \\
& FLa-LLM \citep{chen2024factorized} & --    & --    & --    & --    & --    & 37.13 & 25.12 & 18.38 & 14.20 & 37.25 \\
& $C^{2}$RL \citep{chen2025c}        & --    & --    & --    & --    & --    & 49.32 & 36.28 & 27.54 & 21.61 & 48.21 \\
& Uni-Sign \citep{Li2025sign}       & 55.30 & 42.21 & 32.94 & 26.25 & 56.03 & 55.08 & 42.14 & 32.98 & 26.36 & 56.51 \\
& SignLLM \citep{gong2024llms}      & 42.45 & 26.88 & 17.90 & 12.23 & 39.18 & 39.55 & 28.13 & 20.07 & 15.75 & 39.91 \\
& Sign2GPT \citep{wong2024sign2gpt}  & --    & --    & --    & --    & --    & 41.75 & 28.73 & 20.60 & 15.40 & 42.36 \\
\cmidrule{1-12}
& \textbf{SignDPO} & \textbf{57.87} & \textbf{47.14} & \textbf{35.72} & \textbf{28.95} & \textbf{58.62} & \textbf{57.43} & \textbf{44.24} & \textbf{35.06} & \textbf{28.36} & \textbf{58.55} \\
\bottomrule
\end{tabular}%
}
\end{table*}

\begin{table*}[htbp]
\centering
\caption{SLT results on the OpenASL dataset. Best results are \textbf{bolded}, and \underline{second best} results are underlined.}
\label{tab:openasl_result}
\resizebox{0.7\textwidth}{!}{%
\begin{tabular}{lcccccccccc}
\toprule
\multirow{2}{*}{Method} & \multicolumn{5}{c}{DEV} & \multicolumn{5}{c}{TEST} \\
\cmidrule(r){2-6} \cmidrule(l){7-11}
& B@1 & B@2 & B@3 & B@4 & R@L & B@1 & B@2 & B@3 & B@4 & R@L \\
\midrule
GloFE-VN \citep{lin2023gloss} & 21.06 & 12.34 & 8.68 & 6.68 & 21.37 & 21.56 & 12.74 & 9.05 & 7.06 & 21.75 \\
Conv-GRU \citep{camgoz2018neural} & 16.72 & 8.95 & 6.31 & 4.82 & 16.25 & 16.11 & 8.85 & 6.18 & 4.58 & 16.10 \\
I3D-transformer \citep{shi2022open} & 18.26 & 10.26 & 7.17 & 5.60 & 18.88 & 18.31 & 10.15 & 7.19 & 5.56 & 18.64 \\
OpenASL \citep{shi2022open} & 20.10 & 11.81 & 8.43 & 6.57 & 20.43 & 20.92 & 12.08 & 8.59 & 6.72 & 21.02 \\
Uni-Sign \citep{Li2025sign} & \underline{50.84} & \underline{37.82} & \underline{29.83} & \underline{24.16} & \underline{44.58} & \underline{49.35} & \underline{36.32} & \underline{28.55} & \underline{23.14} & \underline{43.22} \\
$C^{2}$RL \citep{chen2025c} & -- & -- & -- & -- & -- & 31.46 & 21.85 & 16.58 & 13.21 & 31.36 \\
\midrule
\textbf{SignDPO} & \textbf{52.83} & \textbf{38.12} & \textbf{31.80} & \textbf{26.98} & \textbf{46.70} & \textbf{51.44} & \textbf{37.94} & \textbf{29.93} & \textbf{24.35} & \textbf{46.29} \\
\bottomrule
\end{tabular}%
}
\end{table*}

Table \ref{tab:qualitative_openasl} evaluates the models on OpenASL, a dataset characterised by formal, community-orientated discourse. The first example involving a committee recommendation highlights a common SFT failure: semantic weakening. SFT replaces the proactive "encourage" with the passive "want" and collapses the multi-step action "come together and advocate" into a singular "unite," significantly reducing the professional impact of the statement. DPO attempts to rephrase the entire sentence into a descriptive state ("We're a part of..."), which, while fluent, fails to capture the imperative intent of the reference. SignDPO preserves the original "encourage" structure and utilises high-quality semantic substitutes ("gather" and "support"), effectively maintaining the call-to-action tone.

In the second example, which contains formal business terminology, SFT demonstrates imprecise compression, paraphrasing "business agreement" as "business deal" and incorrectly collapsing "certification program" into "getting certified." DPO generates a hallucinated causal link ("happens because it is"), introducing a logical relationship not present in the source. SignDPO maintains the highest structural fidelity, correctly identifying "business" as the subject and using the appropriate relational phrase "would be related to," which mirrors the formal "would be in regards to" of the reference.

The third example, a multi-clause expression of support, illustrates SFT’s struggle with temporal and personal accuracy. SFT shifts the future-tense "will be happy" to the present "are glad," drops the personal pronoun "you" from the support clause, and replaces the specific "concerns" with the vague "what you say." DPO produces a more accurate result but maintains a slightly fragmented structure. SignDPO provides the most elegant and precise translation, condensing the verb phrases ("respond, support, and listen") into a clean, parallel list that preserves every semantic unit of the reference while improving overall naturalness.

Across all three benchmarks, these results confirm that while SFT often simplifies or misinterprets complex prompts and DPO frequently deviates into paraphrastic hallucinations, SignDPO consistently aligns with the reference’s intent. By combining multi-level preference signals, SignDPO effectively captures the linguistic nuances of both Chinese and American sign languages, from colloquial instructions to formal advocacy.

\section{Complete performance results}
Full performance metrics for CSL-Daily (Table~\ref{tab:csl_result}) and OpenASL (Table~\ref{tab:openasl_result}) are presented here to supplement the condensed results in the main paper. These extended versions provide a more granular comparison of all $B@n$ scores across various state-of-the-art methods.

\clearpage
\newpage
\bibliographystyle{acm}
\balance
\bibliography{main} 

\end{document}